\documentclass[letterpaper, 10 pt, conference]{ieeeconf}  

\IEEEoverridecommandlockouts                              

\usepackage{graphicx} 
\usepackage{amsmath} 
\usepackage{amssymb}  
\usepackage{booktabs}
\usepackage[table]{xcolor}
\usepackage{hhline}
\usepackage[nocompress]{cite}

\newif\ifcenterIEEEtablecaption
\makeatletter
\long\def\@makecaption#1#2{%
\ifx\@captype\@IEEEtablestring%
\ifcenterIEEEtablecaption%
\parbox[t]{\hsize}{\footnotesize\centering #1: #2\par}%
\else%
\parbox[t]{\hsize}{\footnotesize\noindent #1: #2}%
\fi%
\@IEEEtablecaptionsepspace%
\else
\@IEEEfigurecaptionsepspace%
\setbox\@tempboxa\hbox{\footnotesize #1.~~ #2}%
\ifdim \wd\@tempboxa >\hsize%
\setbox\@tempboxa\hbox{\footnotesize #1.~~ }%
\parbox[t]{\hsize}{\footnotesize \noindent\unhbox\@tempboxa#2}%
\else%
\ifcenterfigcaptions \hbox to\hsize{\footnotesize\hfil\box\@tempboxa\hfil}%
\else \hbox to\hsize{\footnotesize\box\@tempboxa\hfil}%
\fi\fi\fi}
\makeatother

\title{\LARGE \bf
ReDOA: Real-World Distillation via Observation Alignment for Image-Based Manipulation Policy Learning
}

\author{Kotaro Kajitsuka$^{1,2}$, Masaki Murooka$^{2}$, and Fumio Kanehiro$^{1,2}$
\thanks{*This work was partially supported by JST CREST, Japan, Grant Number JPMJCR2553.}%
\thanks{$^{1}$University of Tsukuba, Tsukuba, Japan.
        Corresponding author: Kotaro Kajitsuka
        ({\tt\small tactile.kajitsuka@aist.go.jp}).}%
\thanks{$^{2}$CNRS-AIST JRL (Joint Robotics Laboratory), IRL3218,
        National Institute of Advanced Industrial Science and Technology (AIST),
        Tsukuba, Japan.}%
}

\begin{document}

\maketitle
\thispagestyle{empty}
\pagestyle{empty}

\begin{abstract}
Data-driven learning for general-purpose robotic manipulation presents a fundamental trade-off between data-collection cost and adaptability to real-world environments.
Imitation learning (IL) requires costly human demonstrations, whereas simulation-based learning is vulnerable to sim-to-real gaps in dynamics and sensor observations.
We propose ReDOA (Real-World Distillation via Observation Alignment), a framework that decouples control learning from perception learning to learn image-based policies in the real world.
A state-based teacher policy trained in simulation is executed on a real robot, and its control knowledge is distilled into an image-based student policy using the collected trajectories.
Observation Alignment resolves observation mismatches between data collection and deployment.
This framework collects successful, efficiently executed trajectories paired with real-world visual observations, without requiring demonstrations from human experts.
In real-robot experiments on three non-prehensile manipulation tasks, including bimanual manipulation, our method achieves better overall performance when jointly considering task success and execution efficiency than both IL based on human demonstrations and policy distillation in simulation.
https://isri-aist.github.io/redoa-project-page/
\end{abstract}

\section{Introduction}
Although demand for automating manipulation tasks in unstructured environments, such as homes and industrial workplaces, continues to grow, successful real-world deployments remain limited.
Data-driven learning has therefore attracted considerable attention as a promising solution.
However, learning both robust control under complex dynamics and reliable perception across diverse real-world environments remains challenging while keeping the costs of collecting demonstrations and setting up environments low.

Existing data-driven approaches face an inherent difficulty in balancing data collection cost and real-world adaptability.
Imitation learning (IL) based on human demonstrations can naturally adapt to real-world visual observations and contact dynamics, but teaching complex dual-arm manipulation through teleoperation requires operator skill and teleoperation interface setup.
Moreover, because vision-based IL is sensitive even to minor appearance variations such as lighting or background shifts, adapting to new environments requires collecting demonstrations again, repeatedly imposing substantial labor and time on human demonstrators.
In contrast, simulation-based methods (sim-to-real approaches) enable training on large datasets at low cost, but gaps in dynamics and sensor observations tend to degrade real-world performance.

%
%
However, recent work has shown that separating control learning in simulation from perception learning in the real world can bypass the sim-to-real gap in observations.
For example, PTLD~\cite{Chen:PTLD:2026} uses real-world policy distillation to bypass the sim-to-real gap of tactile sensors.

\begin{figure}[!t]
  \centerline{\includegraphics[width=0.95\linewidth]{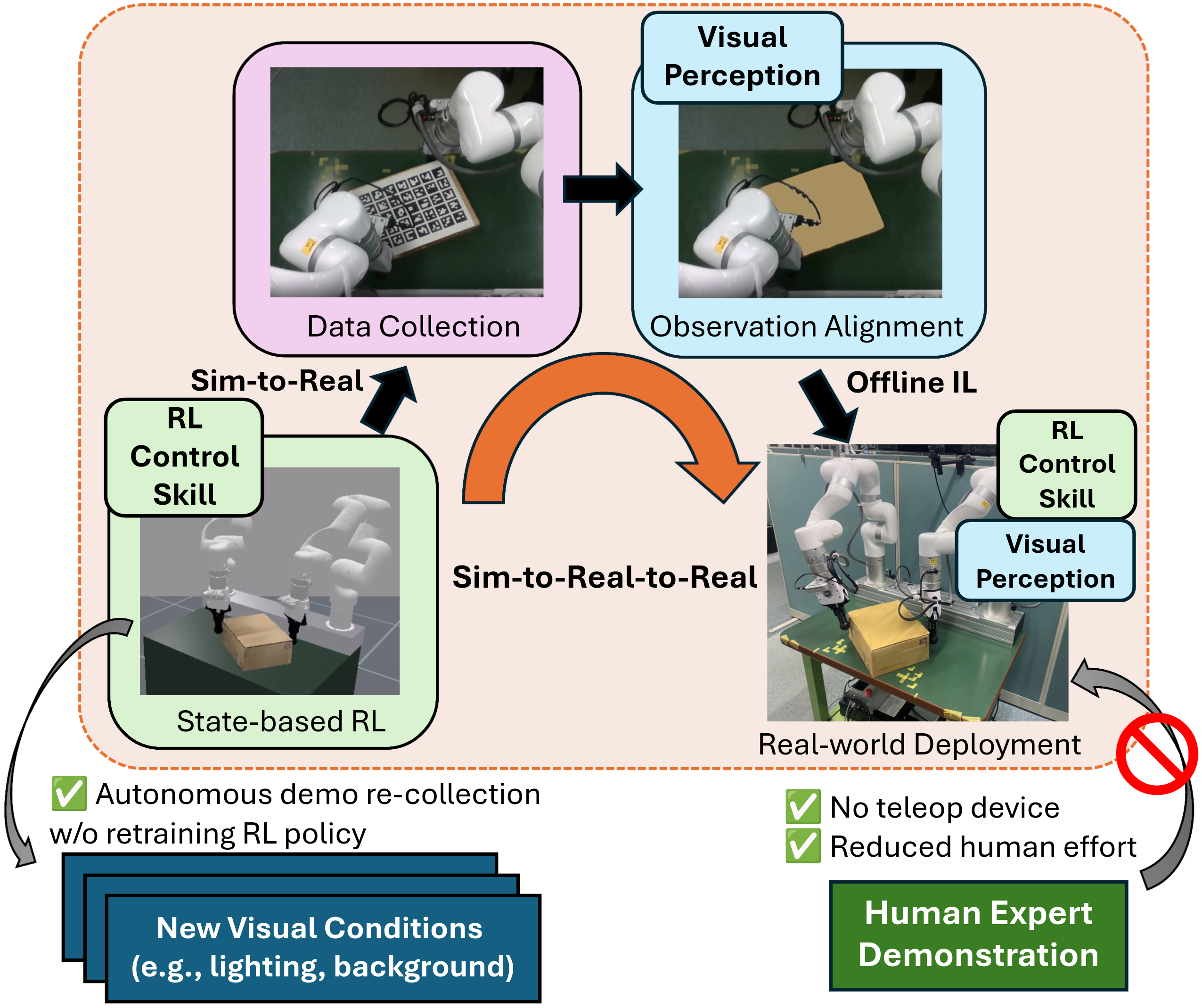}}
  \caption{Conceptual overview of ReDOA (Real-World Distillation via Observation Alignment).}
  \label{fig:concept}
\end{figure}

We propose ReDOA (Real-World Distillation via Observation Alignment), which applies distillation across observation spaces in the real world to image-based manipulation policy learning.
Figure~\ref{fig:concept} illustrates the concept of our method.
The pipeline proceeds as follows.
First, we roll out the teacher policy while estimating the object state using temporarily installed trackers, collecting demonstrations for distillation.
The trackers appear in the collected images; using these images directly to train the student policy creates an observation mismatch when the trackers are removed at deployment.
We therefore perform Observation Alignment to resolve this mismatch and train the student policy using the aligned observations and collected actions.
These stages form a sim-to-real-to-real pipeline.

The framework reduces the burden of human demonstration and bypasses the visual sim-to-real gap by learning perception from images collected in the real world.

We evaluate the proposed method on three non-prehensile manipulation tasks using a real robot.
Non-prehensile manipulation requires continuous visual perception to maintain precise robot--object--environment contact, making it a suitable benchmark.

Prior sim-to-real studies have attempted to limit the performance degradation of vision-based policies by using visual domain randomization~\cite{Josh:VisualDomainRandomization:IROS2017}, green-screening, or texture projection~\cite{Xuanlin:EvalRealinSim:PMLR2025}.
However, these methods suffer from training instability caused by randomization~\cite{Josip:AnalysisRandomization:IROS2022} and require substantial effort to set up the environment.
Our method uses images collected in the real environment to train the student policy, eliminating the need to render its training images in simulation.
Observation Alignment then mitigates the appearance mismatch caused by trackers introduced for real-world data collection.

Unlike conventional IL based on human demonstrations, our method collects data using a consistent policy without relying on human experts or being affected by human fatigue or individual operating habits.
Because the state-based teacher does not depend on visual appearance, it can also autonomously collect new real-world demonstrations without retraining when lighting or backgrounds change.

In summary, our contributions are as follows:
\begin{itemize}
    \item A real-world distillation framework for bypassing the visual sim-to-real gap through Observation Alignment that resolves observation mismatches between demonstration collection and deployment.

    \item Automated generation of high-quality demonstration data without expert teleoperation, reducing human labor costs.

    \item Real-world validation on three non-prehensile tasks, demonstrating the simultaneous achievement of high success rates and high motion efficiency.
\end{itemize}

\section{Related Work}

\begin{figure*}[htbp]
  \centering{\includegraphics[width=0.95\textwidth]{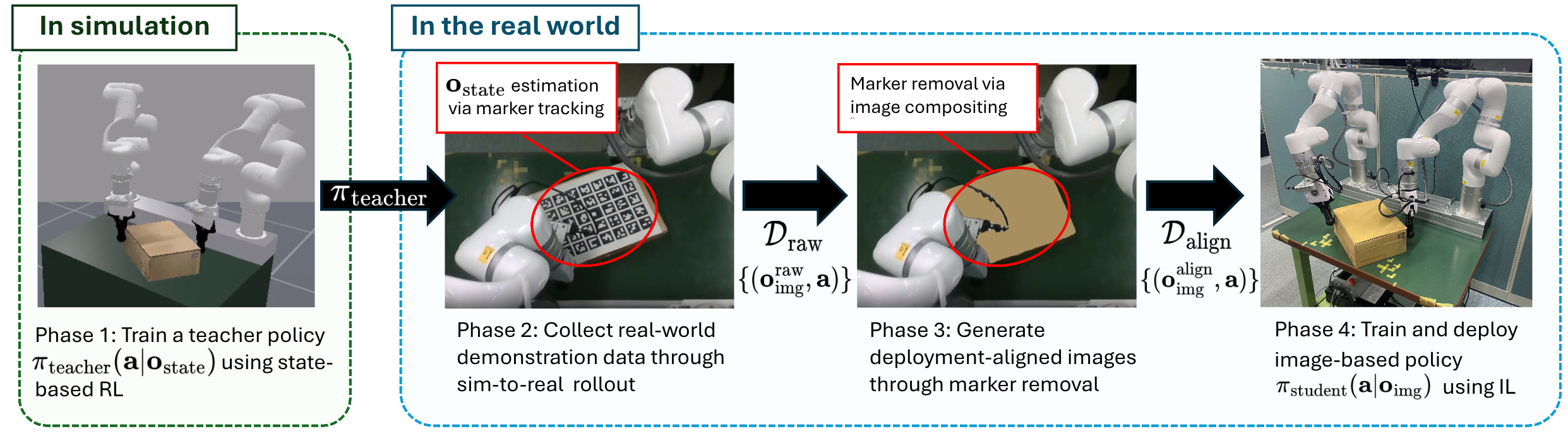}}
  \caption{Overview of the ReDOA (Real-World Distillation via Observation Alignment) learning pipeline.}
  \label{fig:learning_pipline}
\end{figure*}

\subsection{Challenges in IL and Demonstration Data Collection}
\label{subsec:ILDemoData}
Recent advances in IL, including ACT~\cite{Zhao:ACT:RSS2023} and Diffusion Policy~\cite{Chi:DP:RSS2023}, have led to significant success in approaches that generate robot actions directly from visual observations.
However, the performance of these data-driven methods depends strongly on the quality and quantity of demonstration data collected from human experts~\cite{Mandlekar:RoboMimic:2021,Kase:MotorBabbling:JRM2021}.

In particular, the non-prehensile manipulation tasks considered in this study require controlling continuous contact dynamics among the object, the environment, and the robot~\cite{Fabio:NonprehensileSurvey:RAL2018}.
Prior work has proposed RL-based approaches focused primarily on control, including maintaining contact modes during whole-arm manipulation~\cite{Kawarai:WholeArmManipulation:JRM2010}, integrating tactile sensors into the end effectors of a dual-arm robot~\cite{Yijiong:Bi-Touch:RAL2023}, and combining optimization-based demonstration generation with RL~\cite{Shirai:Pivoting:NeurIPS2025}.
These studies show that non-prehensile manipulation requires continuous state observation and sequential action correction to handle delicate physical interactions.
Consequently, collecting a large-scale, high-quality, and consistent real-world dataset for IL through teleoperation is costly.

To address the cost of collecting human demonstrations, RLDG uses reinforcement learning directly on real robots to collect demonstrations containing real-world RGB images and fine-tunes a generalist model on these data~\cite{Xu:RLDG:RSS2025}.
By autonomously generating high-quality trajectories through reward maximization, this approach has been shown to outperform fine-tuning on human demonstrations.

In contrast, this study focuses on training task-specific image-based policies from scratch.

\subsection{Policy Distillation across Observation Spaces}
\label{subsec:related_distill_obs}

Deploying a state-based policy in the real world requires continuously providing the policy with task-relevant state information, such as object poses.
Such state information can be obtained using planar fiducial markers attached to objects~\cite{Cong:SelfAdaptingPushing:IROS2020} or marker-based motion capture~\cite{Sueoka:RealWorldCooperativeTransport:JRM2025}.
However, in settings where many objects of the same type are handled in sequence, attaching markers to each object in advance is impractical.

Image-based policies avoid explicit object-state measurement~\cite{Lerrel:AsymmetricActorCritic:RSS2018,Yarats:DrQ-v2:ICLR2022}.
However, end-to-end policy learning from high-dimensional visual observations requires learning of visual representations and control simultaneously, which tends to reduce sample efficiency and training stability.
In addition, repeatedly rendering visual observations in simulation also increases computational resource requirements and training time.

%
%
%
%
%
%
Policy distillation across observation spaces is used to address these sample-efficiency and computational-cost issues.
A teacher policy is first trained in simulation using privileged observations, and its control knowledge is then distilled, also in simulation, into a student policy that uses observations available at deployment, such as RGB images~\cite{Singh:DextrAH-RGB:2024,He:VIRAL:CVPR2026,Sorokin:Sidewalk:RAL2022}, depth images~\cite{Lum:DextrAH-G:ACRL2024}, or point clouds~\cite{Yamada:Combo-Grasp:PMLR2025,Chen:ClutterDexGrasp:CoRL2025}.

PTLD~\cite{Chen:PTLD:2026} also performs policy distillation in the real world.
It bypasses the sim-to-real gap of tactile sensors by using real-world data to learn to estimate latent representations based on privileged information from tactile and proprioceptive observations.

Our method likewise performs distillation in the real world, using Observation Alignment to resolve the image-observation mismatch caused by the presence of trackers during data collection and their absence at deployment, thereby enabling real-world distillation into an image-based policy.

\subsection{Visual Sim-to-Real Gap}
\label{subsec:related_sim2real_data}
In sim-to-real transfer, the visual gap associated with RGB images is a serious problem and degrades real-robot deployment performance~\cite{Acar:Single-CameraDistill:RAL2024,Liu:FetchBot:2025}.
Real-to-sim-to-real methods such as Re$^3$Sim~\cite{Han:Re3Sim:2025} reduce the visual sim-to-real gap by reconstructing real-world 3D geometry and appearance in photorealistic simulation environments.
Whereas their training data are generated entirely in simulation, our method executes the teacher policy on a real robot to generate demonstrations that directly reflect real-world visual observations and contact dynamics.
Moreover, Re$^3$Sim considers only rigid objects, and its applicability to articulated-object manipulation, as studied here, remains unverified.

By training the student policy using real-world images, our method eliminates the need to transfer from simulated images to real images.
Observation Alignment further mitigates the observation mismatch caused by tracker artifacts in the real data.

\section{Method}
\label{sec:method}
We propose a sim-to-real-to-real pipeline in which a state-based teacher policy trained in simulation is executed on a real robot, and the resulting demonstrations are used to train an image-based student policy.

The underlying problem is formulated as a partially observable Markov decision process (POMDP), denoted by $(\mathcal{S}, \mathcal{O}, \mathcal{A}, \mathcal{T}, R, \gamma)$.
Here, $\mathcal{S}$ denotes the state space, $\mathcal{O}$ the observation space, $\mathcal{A}$ the action space, $\mathcal{T}(\mathbf{s}_{t+1} \mid \mathbf{s}_t, \mathbf{a}_t)$ the state transition function, $R(\mathbf{s}_t, \mathbf{a}_t)$ the reward function, and $\gamma \in [0,1)$ the discount factor.

The proposed learning pipeline is illustrated in Fig.~\ref{fig:learning_pipline}.
The procedure is described in the following four stages.

\subsection{Phase 1: State-based RL in Simulation}
In the first phase, the teacher policy $\pi_\mathrm{teacher}$ is trained in simulation using state-based RL.
Here, the observation space of $\pi_\mathrm{teacher}$ is denoted by $\mathcal{O}_\mathrm{state}$.
In addition to robot proprioception, $\mathcal{O}_\mathrm{state}$ includes pose information about the object.

The objective of the teacher policy is to find an optimal policy $\pi_\mathrm{teacher}^*$ that maximizes the expected discounted return under the transition function $\mathcal{T}_\mathrm{sim}$ of the simulation environment:
\begin{equation}
\pi_\mathrm{teacher}^* = \arg\max_{\pi} \mathbb{E}_{\mathbf{a}_t \sim \pi} \left[ \sum_{t=0}^{T} \gamma^t R(\mathbf{s}_t, \mathbf{a}_t) \right]
\end{equation}

\subsection{Phase 2: Real-world Data Generation with Object Tracking}
In the second phase, the teacher policy $\pi_\mathrm{teacher}$, trained in simulation, is executed on the real robot to autonomously collect demonstration data for training the student policy.

During real-world rollouts, trackers whose poses can be easily estimated are placed on the object and tracked online.
The estimation function $\phi$ reconstructs the state from the image observation $\mathbf{o}_\mathrm{img}^\mathrm{raw}$, in which the trackers are visible, as follows:
\begin{equation}
\mathbf{o}_\mathrm{state} = \phi(\mathbf{o}_\mathrm{img}^\mathrm{raw})
\end{equation}

Throughout this paper, $\mathbf{o}_\mathrm{img}$ denotes an observation comprising an RGB image and robot proprioception, such as joint angles.

Actions $\mathbf{a}$ are generated from this reliable state observation $\mathbf{o}_\mathrm{state} \in \mathcal{O}_\mathrm{state}$, and demonstration data are collected accordingly.
Formally, we write
\begin{equation}
\mathbf{a} \sim \pi_\mathrm{teacher}(\cdot|\mathbf{o}_\mathrm{state})
\end{equation}
\begin{equation}
\mathcal{D}_\mathrm{raw} := \{(\mathbf{o}_\mathrm{img}^\mathrm{raw}, \mathbf{a})\}\text{.}
\end{equation}

\subsection{Phase 3: Observation Alignment for Distillation}

In the third phase, we perform tracker artifact removal on the observations in $\mathcal{D}_\mathrm{raw}$ to construct a deployment-aligned demonstration dataset for offline IL.

The raw observations $\mathbf{o}_\mathrm{img}^\mathrm{raw}$ contain tracker artifacts that are absent at deployment.

We define Observation Alignment as a transformation $f$ that maps data-collection images to the deployment observation domain.
In this study, we implement this transformation through image compositing, which fills the regions occluded by the trackers.
The resulting deployment-aligned image is given by
\[
\mathbf{o}_\mathrm{img}^\mathrm{align} = f(\mathbf{o}_\mathrm{img}^\mathrm{raw})
\]

We use planar fiducial markers as trackers.
We chose them because they provide stable estimates of the object's six-degree-of-freedom pose.
In addition, a marker board with multiple markers printed at equal intervals allows pose estimation to continue from the remaining markers even when some are occluded by the robot.

\begin{figure}[!t]
  \centerline{\includegraphics[width=0.95\linewidth]{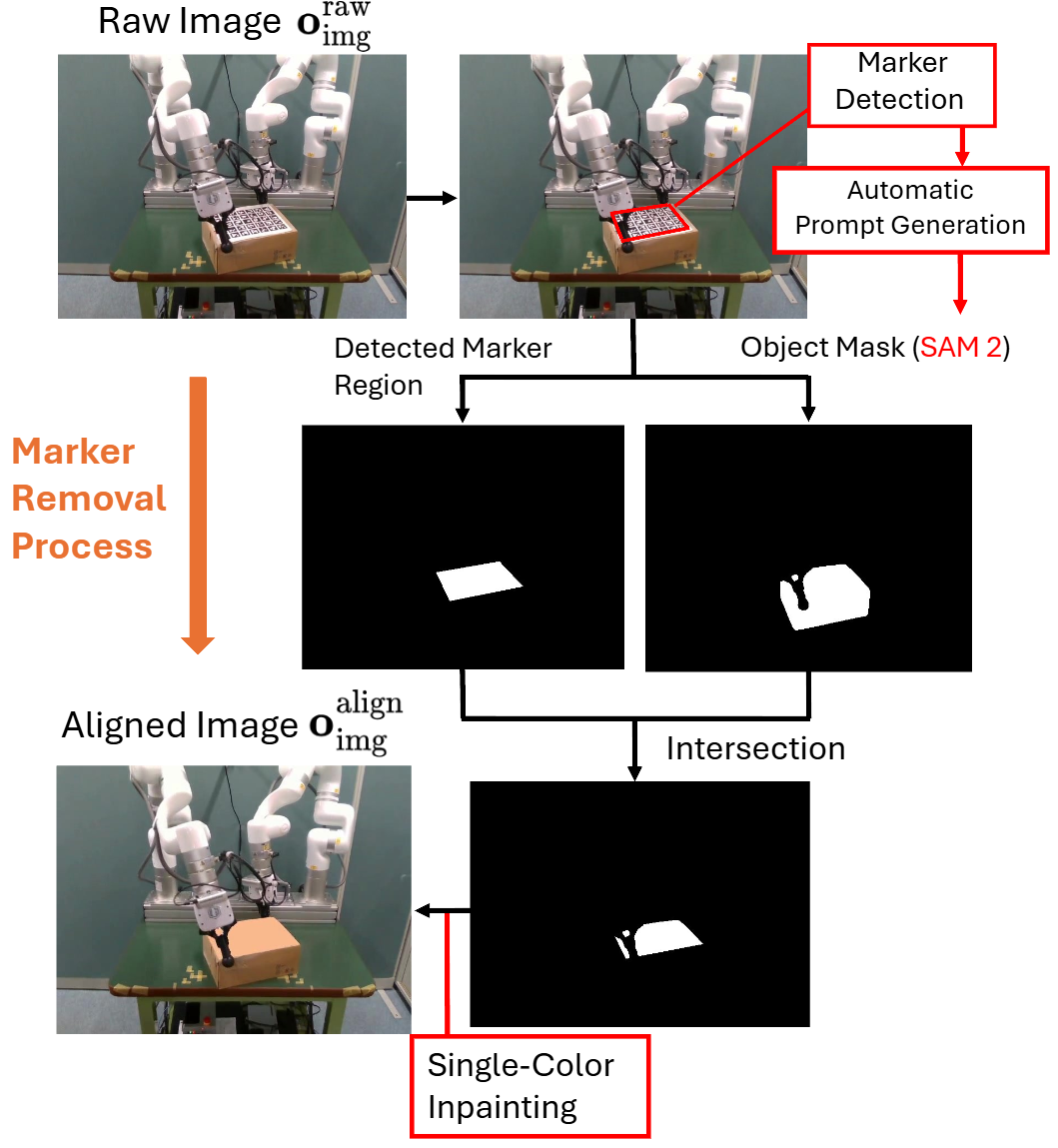}}
  \caption{Workflow for tracker artifact removal from raw demonstration images.
  We used planar fiducial markers as trackers for this whole study.}
  \label{fig:MarkerRemovalPipeline}
\end{figure}

The overall workflow is illustrated in Fig.~\ref{fig:MarkerRemovalPipeline}.
The specific procedure is as follows.

\begin{enumerate}
    \item The marker region is detected in the image using an existing marker detector.
    \item The target object is segmented using Segment Anything Model~2 (SAM~2)~\cite{Nikhila:SAM2:2024} to distinguish its pixels within the detected marker region from pixels belonging to the robot or other objects.
    In this process, the detected marker region is used to provide SAM~2 with a positive prompt for the target object.
    \item The intersection between the marker region and the segmented object region is computed, and this region is identified as the mask.
    \item The identified masked region is inpainted using the original color values of the object.
\end{enumerate}

\medskip

This marker-removal process makes it possible to inpaint only the marker regions while preserving image regions containing the robot, even when some markers are occluded by the robot or other objects.
Applying the tracker artifact removal process $f$ to the raw dataset $\mathcal{D}_\mathrm{raw}$ yields the deployment-aligned demonstration dataset
\[
\mathcal{D}_\mathrm{align} = \{(f(\mathbf{o}_\mathrm{img}^\mathrm{raw}), \mathbf{a})\} = \{(\mathbf{o}_\mathrm{img}^\mathrm{align}, \mathbf{a})\}.
\]

\subsection{Offline IL}

Offline IL is performed using the deployment-aligned demonstration dataset $\mathcal{D}_\mathrm{align}$.
The objective of the student policy $\pi_{\mathrm{student}}$ is to approximate the expert action distribution, and the training objective is defined as
\begin{equation}
\pi_\mathrm{student}^* = \arg\min_{\pi} \mathbb{E}_{(\mathbf{o}_\mathrm{img}^\mathrm{align}, \mathbf{a}) \sim \mathcal{D}_\mathrm{align}}
 \left[
        -\log \pi(\mathbf{a}|\mathbf{o}_\mathrm{img}^\mathrm{align})
\right].
\end{equation}

Our method uses image compositing to align images with deployment observation conditions, then trains the student policy using the aligned observations and teacher actions.

\section{Experimental Setup}
\subsection{Hardware and Task Details}
We used a dual-arm platform comprising two UFACTORY xArm7 manipulators.
For each task, we placed one fixed external camera (RealSense D435) at a position that provided a clear view of the workspace.

\begin{figure}[!t]
  \centering
  \begin{minipage}[t]{0.48\linewidth}
    \centering
    \includegraphics[width=\linewidth]{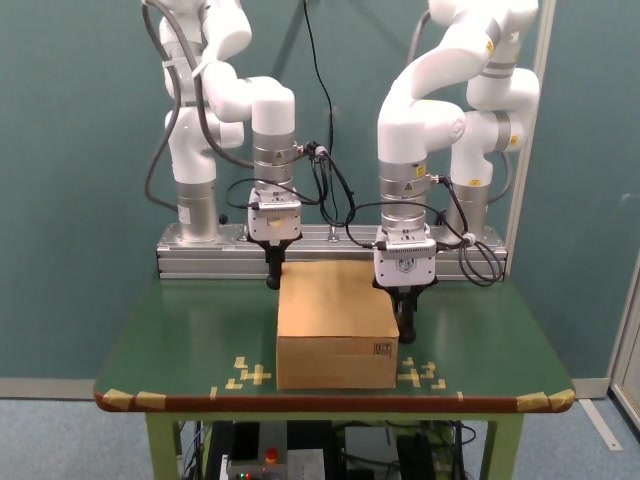}\par
    {\scriptsize (a) Box Rotation Task\par}
  \end{minipage}
  \hfill
  \begin{minipage}[t]{0.48\linewidth}
    \centering
    \includegraphics[width=\linewidth]{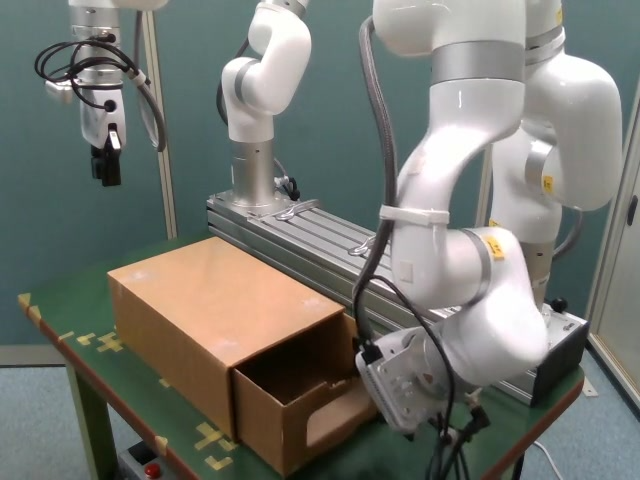}\par
    {\scriptsize (b) Drawer Opening Task\par}
  \end{minipage}

  \vspace{1mm}

  \begin{minipage}[t]{0.48\linewidth}
    \centering
    \includegraphics[width=\linewidth]{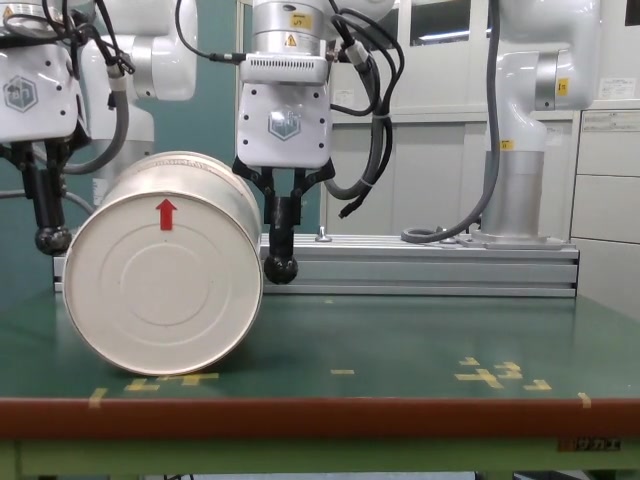}\par
    {\scriptsize (c) Cylinder Rolling Task\par}
  \end{minipage}
  \caption{Real-world manipulation tasks.}
  \label{fig:experimental_tasks}
\end{figure}

The experimental tasks consisted of the Box Rotation Task, the Drawer Opening Task, and the Cylinder Rolling Task, as shown in Fig.~\ref{fig:experimental_tasks}.

In the Box Rotation Task, the robot rotates a box from a randomly assigned initial pose to 90~degrees, where the box is perpendicular to the silver pedestal.
A trial is successful if the box comes to rest at an orientation within 5.0~degrees of 90.0~degrees.
For evaluation, we defined 10 initial box poses relative to the center of the worktable.
The poses were selected from within ranges of $\pm80$~mm in lateral displacement, $-30$ to $+50$~mm in longitudinal displacement, and $\pm25$~degrees in orientation.
Each trial had a time limit of 30.0~s.

In the Drawer Opening Task, the robot opens a closed drawer from a randomly assigned initial position by at least 10~cm.
Because the proposed method obtains state information from object-mounted trackers, it can be readily applied to tasks involving articulated objects by mounting trackers on their individual links.
For evaluation, we similarly defined 10 initial drawer positions relative to the reference point.
The positions were selected from within ranges of $\pm45$~mm in lateral displacement and $\pm30$~mm in longitudinal displacement.
Each trial had a time limit of 7.5~s.

In the Cylinder Rolling Task, the robot rolls a tapered cylindrical object until it comes to rest with its label facing upward, corresponding to an orientation of 0~degrees.
Because the object can quickly reach high speeds while rolling, we expected demonstration collection via teleoperation to be difficult and selected this task to evaluate the effectiveness of the proposed method.
For evaluation, we similarly defined 10 initial cylinder poses relative to the reference point.
The poses were selected from within ranges of $\pm120$~mm in lateral displacement, $-30$ to $+60$~mm in longitudinal displacement, and $+90$ to $+270$~degrees in orientation.
Each trial had a time limit of 15~s.

For each task, the 10 initial configurations used for evaluation intentionally included two or three configurations outside the initial-pose distribution used during demonstration collection.
We treated these as out-of-distribution (OOD) initial configurations and used them to evaluate policy generalization beyond the training distribution.

Across all tasks, the robot was configured to stop automatically when the contact force exceeded a predefined threshold, and any such contact was considered a failure.

Accordingly, we used the following three metrics to evaluate the proposed method in terms of both task-completion capability and motion efficiency.
Here, $\uparrow$ indicates higher is better, and $\downarrow$ indicates lower is better.

\begin{itemize}
    \item Success Rate (SR$\uparrow$)
    \item Average Time for Successful Rollouts (ATR$\downarrow$)
    \item Throughput-with-Regret (TPR$\uparrow$)
\end{itemize}

TPR$\uparrow$ is a metric proposed in~\cite{Ranawaka:SAIL:PMLR2025} that combines success rate and execution time for successful rollouts.
It is defined as follows:
\begin{equation}
    \text{TPR} = \frac{1}{N} \sum_{i=1}^{N} \left( \left( \frac{1}{t_i} \cdot S_i \right) - \left( \frac{1}{t^{\max}} \cdot (1 - S_i) \right) \right)\text{,}
\end{equation}
where $t_i$ denotes the execution time of trial $i$, $t^{\max}$ is the maximum time limit allowed for each trial, and $S_i \in \{0,1\}$ indicates the success of trial $i$.
It assigns the same penalty to every failure while rewarding faster successes more highly, enabling a joint assessment of success rate and execution time.

In the evaluation experiments, each policy trained with a given random seed was tested on the 10 initial configurations defined for evaluation.

\subsection{Implementation Details of the Proposed Method}
\noindent\textbf{State-Based RL in Simulation} \quad
In all tasks, each action specified incremental joint position commands.
The state observation included a task-specific subset of the following variables: robot joint positions and velocities, current and target tool-center-point (TCP) poses, and current and previous object poses.
We used both arms for Box Rotation and Cylinder Rolling and a single arm equipped with a gripper for Drawer Opening.
During training, we added noise to the object-pose observations to reduce the sim-to-real gap and stabilize real-world execution.

\smallskip
For each task, the dense reward comprised four components: (1) progress toward the target object state, (2) end-effector proximity to the desired contact points, (3) end-effector retreat after task completion, and (4) penalties for undesirable contacts such as self-collision.
This reward design was informed by prior work, including ManiSkill~\cite{Siyunan:Maniskill3:RSS2025} and Bi-Touch~\cite{Yijiong:Bi-Touch:RAL2023}.
We used Soft Actor-Critic (SAC)~\cite{Haarnoja:SAC_Maniskill:PMLR2018} for Box Rotation and Proximal Policy Optimization (PPO)~\cite{Schulman:PPO:arXiv2017} for the other two tasks.
We trained all RL policies in ManiSkill at a control frequency of 20~Hz.

\medskip
\noindent\textbf{Demonstration Collection and Tracker Artifact Removal} \quad
We used the RoboManipBaselines software framework~\cite{Murooka:RoboManipBaselines:IEEEAccess2026} for demonstration collection and subsequent IL.
We used an ArUco Marker Board~\cite{Sergio:ArUcoMarker:PR2014} to track the manipulated object.

The number of collected episodes was determined by the initial configurations and data-collection constraints of each task.
During data collection for all tasks, we repeated rollouts following teacher-policy failures.
For Box Rotation, the grid of initial configurations yielded 45 episodes.
For this task, we used two teacher policies because we did not obtain a single policy that consistently succeeded across the collection region.
For Drawer Opening, the real-world teacher policy failed to complete the task from some initial positions, preventing the collection of successful trajectories.
Possible causes include the sim-to-real gap in the controller.
We therefore reduced the real-world collection area and used 40 episodes collected within the region where the teacher succeeded.
For Cylinder Rolling, the markers left the camera's field of view during execution from some initial configurations; we excluded these configurations and used 36 episodes.
The student policy input for IL was an observation $\mathbf{o}_\mathrm{img}^\mathrm{align}$ comprising a single marker-removed RGB image from the fixed external camera and robot joint angles.

\medskip
\noindent\textbf{Imitation Learning} \quad
We used Diffusion Policy (DP)~\cite{Chi:DP:RSS2023}, a strong and widely adopted imitation learning algorithm.
Visual observations had a resolution of $320 \times 240$ pixels and were cropped to $288 \times 216$ pixels before being input to the network.
We used a CNN backbone and trained the policy for 1000 epochs.
During deployment of DP, the control frequency was approximately 20~Hz on average.

\begin{figure*}[!t]
  \centering{\includegraphics[width=0.95\textwidth]{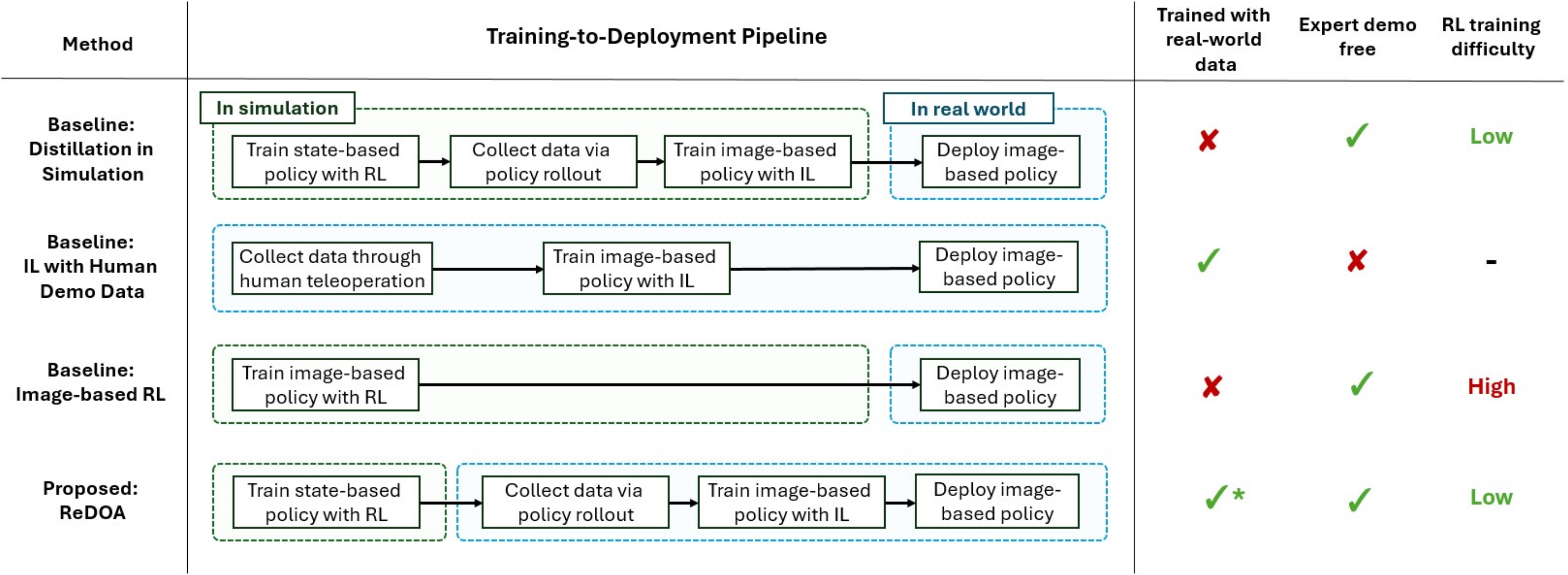}}
  \caption{
Comparison of the training pipelines of the proposed method and the three baseline methods.
The key characteristics of each method are also summarized on the right.\\
* Real-world data are used; however, the training images undergo tracker artifact removal.
  }
  \label{fig:ComparedBaselines}
\end{figure*}

\section{Experimental Results}
\subsection{Overview of Compared Methods}

Fig.~\ref{fig:ComparedBaselines} summarizes the workflows and characteristics of the learning pipelines for the three baseline methods and the proposed method.

\textbf{Distillation in Simulation} is a method that collects demonstration data from rollouts of state-based RL in simulation and then trains an image-based policy on the collected data via offline IL.
Because the entire data collection process is completed in simulation, this method enables efficient large-scale data generation.
On the other hand, the method is more susceptible to visual and physical sim-to-real gaps when deployed in the real world.
For this baseline, we randomized the lighting conditions in simulation and sampled the initial object configurations more densely, resulting in a training dataset with twice as many demonstration episodes as those used for the other methods.
To examine the effect of the environment in which demonstrations are generated, we used the same offline IL procedure (Diffusion Policy) for this baseline and the proposed method.

\textbf{IL with Human Demonstrations} is a representative approach that directly uses real-world data.
A skilled teleoperator collected demonstrations using two 3Dconnexion SpaceMouse devices.
We trained a Diffusion Policy using a dataset containing the same number of episodes as the proposed method and evaluated its real-robot performance.
This baseline provides a reference for IL performance when demonstrations from a skilled human operator are available.

\textbf{Image-based RL} is an approach that directly trains a policy with RL using image observations as input.
Although this approach can be conducted entirely in simulation, it has the disadvantage of being challenging due to training instability and high computational cost.

\textbf{ReDOA (Ours)} is described in detail in Section~\ref{sec:method}.

\begin{table*}[t]
\centering
\centerIEEEtablecaptiontrue
\caption{Performance evaluation results.
Results are averaged over 5 random seeds (10 trials per seed, 50 trials in total).}
\centerIEEEtablecaptionfalse
\label{tab:RealEvalMultipleTasks}
{\renewcommand{\arraystretch}{1.15}
\resizebox{\textwidth}{!}{
\begin{tabular}{l|ccc|ccc|ccc}
\specialrule{\heavyrulewidth}{0pt}{0pt}
&
\multicolumn{3}{c|}{Box Rotation}
&
\multicolumn{3}{c|}{Drawer Opening}
&
\multicolumn{3}{c}{Cylinder Rolling}
\\
\hhline{~|---|---|---}
\raisebox{0.8ex}{Method}
& SR (\%)$\uparrow$
& ATR (s)$\downarrow$
& TPR$\uparrow$
& SR (\%)$\uparrow$
& ATR (s)$\downarrow$
& TPR$\uparrow$
& SR (\%)$\uparrow$
& ATR (s)$\downarrow$
& TPR$\uparrow$
\\
\specialrule{\lightrulewidth}{0pt}{0pt}
Distill. in Sim
& 14.0 & \textbf{7.56} & -0.010
& - & - & -
& - & - & -
\\

IL with Human Demo
& \textbf{76.0} & 20.26 & 0.037
& \textbf{78.0} & 5.49 & 0.119
& \textbf{72.0} & 9.84 & 0.056
\\

\textbf{ReDOA (Ours)}
& 70.0 & 8.81 & \textbf{0.075}
& 64.0 & \textbf{2.75} & \textbf{0.187}
& 68.0 & \textbf{8.03} & \textbf{0.068}
\\
\specialrule{\heavyrulewidth}{0pt}{0pt}
\end{tabular}
}
}
\end{table*}

\subsection{Real-World Comparison with Baselines}
We compare the proposed method with Distillation in Simulation and IL with Human Demonstrations through real-world experiments.
The former evaluates whether performing policy distillation in the real world alleviates the visual sim-to-real gap, while the latter assesses how effectively the proposed framework can replace expert human demonstrations.
Since Image-based RL was evaluated only in simulation, we discuss it separately in Section~\ref{subsec:image_based_rl_training}, focusing on its training behavior rather than real-world performance.
As the primary evaluation metric, we use Throughput-with-Regret (TPR$\uparrow$), which jointly captures task success and execution efficiency.

The quantitative results are summarized in Table~\ref{tab:RealEvalMultipleTasks}.
The proposed method achieved the highest TPR$\uparrow$ on all three tasks.

The comparison with Distillation in Simulation highlights the importance of performing policy distillation in the real world.
The baseline achieved only 14.0\% SR$\uparrow$ on Box Rotation, and therefore was not evaluated on the remaining tasks.
Despite using randomized lighting conditions and twice as many demonstrations as the other methods, the policy did not achieve sufficient real-world transfer performance under the dataset size and training conditions used in this experiment.
Furthermore, successful trials were limited to relatively easy initial configurations in which the box was already rotated by approximately 15~degrees.
Consequently, its Average Time for Successful Rollouts (ATR$\downarrow$), computed only over successful trials, was the shortest (7.56 s).

The proposed method also outperformed IL with Human Demonstrations in execution efficiency.
Averaged over all three tasks, it reduced ATR$\downarrow$ by 5.33 s, with the largest improvement (11.45 s) observed on Box Rotation.
This difference primarily reflects the execution efficiency of the demonstration trajectories themselves: the average ATR$\downarrow$ of the RL-generated and human demonstrations was 6.88 s and 21.98 s, respectively.

The efficiency gain was particularly pronounced in Box Rotation, where collecting demonstrations through teleoperation is inherently challenging.
During manipulation, the box frequently occluded one of the end-effectors, preventing the operator from continuously monitoring the contact relationships between the end-effectors, the object, and the tabletop.
As a result, the operator tended to execute conservative motions with larger safety margins.
In contrast, the RL teacher policy had already learned efficient manipulation strategies in simulation based on the spatial relationship between the robot and the object.
Consequently, it could execute the task efficiently using observations such as the object pose and robot joint states.
These findings suggest that the proposed framework can generate efficient behaviors that are difficult to demonstrate through vision-based teleoperation.

The proposed method achieved an average SR$\uparrow$ that was 8.0 percentage points lower than IL with Human Demonstrations.
We attribute this gap to two factors.
First, unlike the human demonstrations, which maintained relatively large safety margins around the manipulated object, the RL-generated demonstrations occasionally contained trajectories passing closer to the object.
As a result, for some out-of-distribution initial configurations, unintended collisions reduced the success rates in Box Rotation and Drawer Opening.
This observation suggests a trade-off between execution efficiency and robustness to out-of-distribution initial states.

Second, the marker-removal preprocessing may have slightly degraded the visual observations.
Filling the tracker regions with a uniform color removed local shadow cues and introduced subtle appearance discontinuities, which may have negatively affected the learned policy.
We investigate this effect in the following ablation study.

Overall, although the proposed method achieved a lower SR$\uparrow$ than IL with Human Demonstrations, it achieved the highest TPR$\uparrow$ across all evaluated tasks.
These results indicate that the proposed framework attains strong task performance and efficient execution while significantly reducing the need for costly human data collection.

\begin{table}[b]
\centering
\caption{Ablation study evaluating the effect of marker removal.
Results are averaged over 3 random seeds (10 trials per seed, 30 trials in total).}
\label{tab:results2}
\resizebox{\columnwidth}{!}{
\begin{tabular}{ccccc}
\toprule
Image Compositing & Deploy Env & SR (\%)$\uparrow$ & ATR (s)$\downarrow$ & TPR$\uparrow$ \\
\midrule
Proposed & Marker-Free & 70.0 & 8.62 & 0.079 \\
w/o Image Compositing & Marker-Free & 53.3 & 8.94 & 0.047 \\
w/o Image Compositing & Marker-Based & 73.3 & 7.95 & 0.086 \\
\bottomrule
\end{tabular}
}
\end{table}

\subsection{Ablation Study on Marker Removal}

The proposed method removes the visible trackers from the demonstration images by filling their regions with a uniform color.
This process removes the original shadows and creates slight color discontinuities between the filled and surrounding regions.
If the IL policy learns to rely on these artificial appearance cues, its performance may degrade at deployment.

To evaluate this effect, we conducted an ablation study on the Box Rotation Task by varying whether image compositing was applied and whether the deployment environment contained markers, as shown in Table~\ref{tab:results2}.

The bottom row reports the results of deploying the policy trained without image compositing in the Marker-Based environment.
This setting requires markers during deployment but does not require their removal from the training images, thereby matching the visual conditions between training and deployment.
We therefore use this result as an empirical upper-bound reference for image-based policy performance in our problem setting.

When the same policy was deployed directly in the Marker-Free environment, as shown in the second row, SR decreased by 20.0 percentage points, from 73.3\% to 53.3\%, and TPR decreased from 0.086 to 0.047.
This result shows that mismatched marker conditions between training and deployment substantially degrade policy performance.

With image compositing, the proposed method achieved an SR of 70.0\% and a TPR of 0.079 in the same Marker-Free environment, differing from the upper-bound reference by 3.3 percentage points and 0.007, respectively.
Thus, image compositing enabled near-upper-bound performance in a marker-free deployment environment.

\subsection{Training Behavior of Image-Based RL}
\label{subsec:image_based_rl_training}

\begin{figure}[!t]
  \centerline{\includegraphics[width=0.95\linewidth]{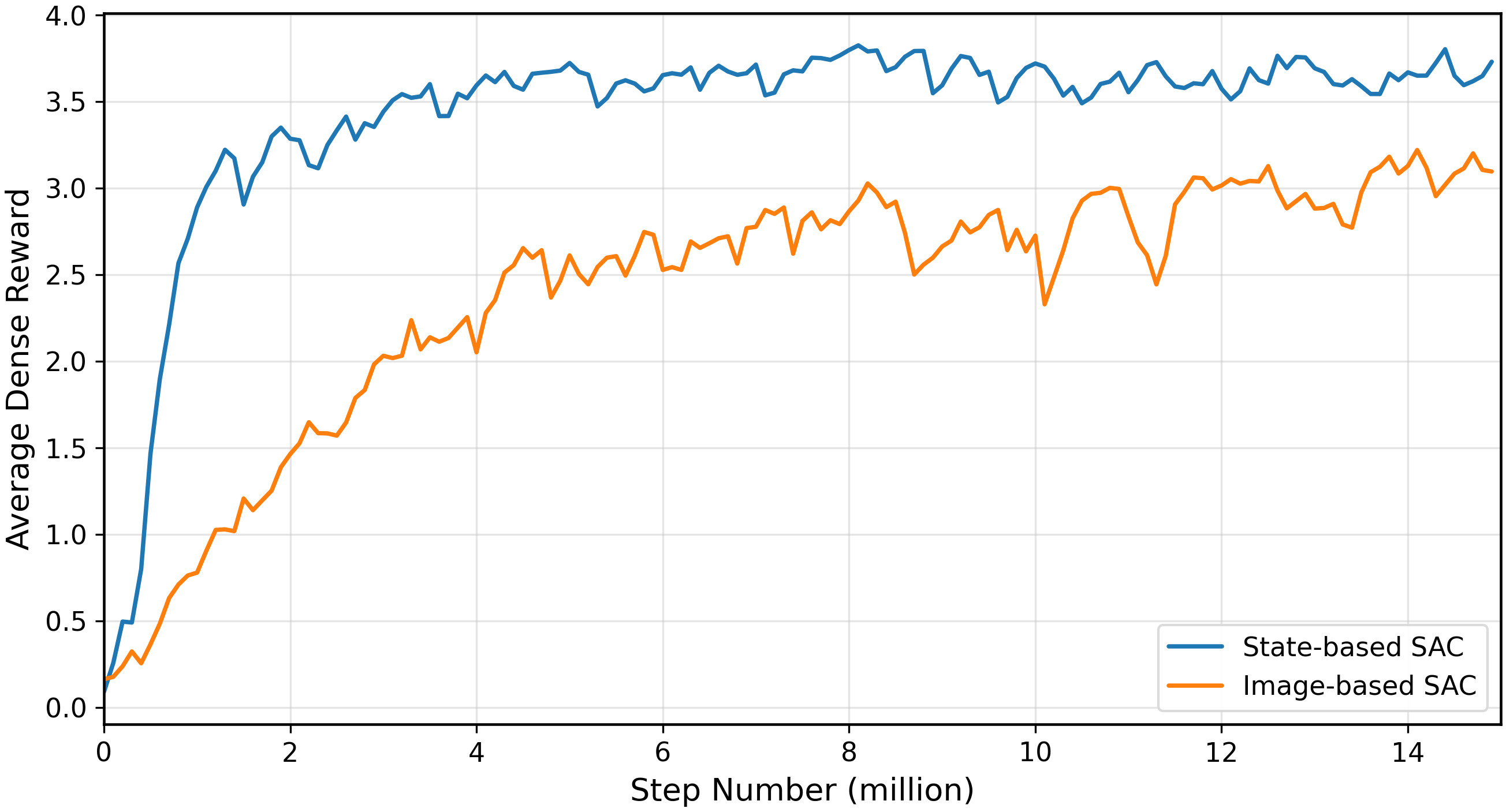}}
  \caption{Training curves of state-based SAC and image-based SAC in simulation.
  Curves are smoothed for visualization.
  State-based SAC learns substantially faster and converges to a higher return than image-based SAC.}
  \label{fig:DenseRewardGraph}
\end{figure}

Figure~\ref{fig:DenseRewardGraph} shows the simulated training curves of state-based SAC and image-based SAC for Box Rotation.
While state-based SAC converged to a high return at an early stage, image-based SAC also made learning progress but remained inferior to state-based SAC in both convergence speed and final performance.
In addition, because image-based SAC involves image rendering in simulation, its computational cost was higher than that of state-based SAC.
In practice, training for 15M timesteps required 12 hours for state-based SAC and 26 hours for image-based SAC.

Thus, although image-based RL can serve as an end-to-end solution for this task, it still faces challenges in both training efficiency and achievable performance.

\section{Limitations}

The first and most fundamental limitation of our method is its assumption that an RL teacher policy capable of completing the target task on the real robot can be obtained in advance.
Consequently, the method is difficult to apply to objects whose dynamics cannot be modeled with sufficient fidelity in simulation, such as cardboard with folds.
It is also difficult to apply the method to objects such as cloth, for which the state information cannot be measured reliably in the real world.
Extending the method to such objects will require advances in simulation fidelity and real-world state-estimation techniques.

A second limitation is that the current marker-removal process fills marker regions with a uniform color and therefore cannot reproduce shadows, brightness variations, or printed patterns, leaving appearance inconsistencies with the surrounding regions.
Aligning the filled regions with their surroundings using scanned texture projection or image harmonization~\cite{Tsai:DeepImageHarmonization:CVP2017} could improve IL performance and broaden the method's applicability to objects with text or graphical patterns.

A third limitation is that using planar fiducial markers as trackers restricts the method to objects with rigid planar surfaces and prevents it from handling occlusions that hide all markers.
Embedding a magnetic tracker inside an object entails substantial setup costs, but this approach can overcome these limitations, thereby extending the method to a wider range of objects and tasks.
Because the tracker does not appear in the images, this approach can also eliminate the need for tracker removal.

Our evaluation is limited to offline IL using fixed datasets.
Introducing DAgger-style data aggregation into ReDOA could mitigate state-distribution mismatch and further improve performance by learning additional teacher actions for states visited by the student policy.
Evaluating the benefit of this extension is left for future work.
The simulation-distillation baseline could also benefit from DAgger-style learning.
Since we have not evaluated this effect, the present comparison does not establish ReDOA's superiority over DAgger-style distillation in simulation.

\section{Conclusion}

%
In this paper, we introduced ReDOA, a real-world distillation framework for learning image-based manipulation policies while bypassing the visual sim-to-real gap.
We roll out a state-based teacher trained in simulation on a real robot, fill tracker regions in the collected images through image compositing, and train a student policy.
The key idea is to mitigate the mismatch between data-collection and deployment images caused by the trackers used to measure the teacher's state observations, transferring control knowledge to a policy that requires no trackers during deployment.

Experiments on three non-prehensile manipulation tasks, including bimanual tasks, showed success rates close to those of IL based on human demonstrations and higher motion efficiency, demonstrating a balance between task success and execution efficiency.
The ablation study further confirmed that tracker removal, our implementation of Observation Alignment, reduces the observation mismatch between data collection and deployment and is an important component supporting performance.
These results demonstrate that ReDOA is a practical pipeline for learning robust, motion-efficient image-based policies.









\bibliographystyle{IEEEtran}
\bibliography{IEEEabrv,root}

@ARTICLE{Han:Re3Sim:2025,
  author  = {Xiaoshen Han and Junqiu Yu and Minghuan Liu and Yilun Chen and Xiaoyang Lyu and Yang Tian and Bolun Wang and Weinan Zhang and Jiangmiao Pang},
  title   = {{Re$^3$Sim}: Generating High-Fidelity Simulation Data via 3D-Photorealistic Real-to-Sim for Robotic Manipulation},
  journal = {arXiv preprint arXiv:2502.08645},
  year    = {2025}
}

@INPROCEEDINGS{Lerrel:AsymmetricActorCritic:RSS2018,
    AUTHOR    = {Lerrel Pinto AND Marcin Andrychowicz AND Peter Welinder AND Wojciech Zaremba AND Pieter Abbeel}, 
    TITLE     = {Asymmetric Actor Critic for Image-Based Robot Learning}, 
    BOOKTITLE = {Proceedings of Robotics: Science and Systems (RSS)}, 
    YEAR      = {2018}, 
    DOI       = {10.15607/RSS.2018.XIV.008} 
}

@ARTICLE{Acar:Single-CameraDistill:RAL2024,
  author={Acar, Cihan and Binici, Kuluhan and Tekirdağ, Alp and Wu, Yan},
  journal={IEEE Robotics and Automation Letters}, 
  title={Visual-Policy Learning Through Multi-Camera View to Single-Camera View Knowledge Distillation for Robot Manipulation Tasks}, 
  year={2024},
  volume={9},
  number={1},
  pages={691-698},
  doi={10.1109/LRA.2023.3336245}}

@ARTICLE{Murooka:RoboManipBaselines:IEEEAccess2026,
  author={Murooka, Masaki and Motoda, Tomohiro and Nakajo, Ryoichi and Oh, Hanbit and Makihara, Koshi and Shirai, Keisuke and Ogata, Tetsuya and Domae, Yukiyasu},
  journal={IEEE Access}, 
  title={{RoboManipBaselines}: A Unified Framework for Imitation Learning in Robotic Manipulation Across Real and Simulation Environments},
  year={2026},
  volume={14},
  number={},
  pages={97896-97909},
  doi={10.1109/ACCESS.2026.3691054}}

@INPROCEEDINGS{Yamada:Combo-Grasp:PMLR2025,
  title = 	 {{COMBO-Grasp}: Learning Constraint-Based Manipulation for Bimanual Occluded Grasping},
  author =       {Yamada, Jun and Mitchell, Alexander Luis and Collins, Jack and Posner, Ingmar},
  booktitle = 	 {Conference on Robot Learning (CoRL)},
  pages = 	 {1102--1119},
  year = 	 {2025},
  volume = 	 {305},
  publisher =    {PMLR}
}

@INPROCEEDINGS{Xu:RLDG:RSS2025, 
    AUTHOR    = {Charles Xu AND Qiyang Li AND Jianlan Luo AND Sergey Levine}, 
    TITLE     = {{RLDG: Robotic Generalist Policy Distillation via Reinforcement Learning}}, 
    BOOKTITLE = {Proceedings of Robotics: Science and Systems (RSS)}, 
    YEAR      = {2025}, 
    DOI       = {10.15607/RSS.2025.XXI.028} 
}

@ARTICLE{Yijiong:Bi-Touch:RAL2023,
  author={Lin, Yijiong and Church, Alex and Yang, Max and Li, Haoran and Lloyd, John and Zhang, Dandan and Lepora, Nathan F.},
  journal={IEEE Robotics and Automation Letters}, 
  title={{Bi-Touch}: Bimanual Tactile Manipulation With Sim-to-Real Deep Reinforcement Learning},
  year={2023},
  volume={8},
  number={9},
  pages={5472-5479},
  doi={10.1109/LRA.2023.3295991}}

@INPROCEEDINGS{Cong:SelfAdaptingPushing:IROS2020,
  author={Cong, Lin and Görner, Michael and Ruppel, Philipp and Liang, Hongzhuo and Hendrich, Norman and Zhang, Jianwei},
  booktitle={2020 IEEE/RSJ International Conference on Intelligent Robots and Systems (IROS)},
  title={Self-Adapting Recurrent Models for Object Pushing from Learning in Simulation},
  year={2020},
  pages={5304--5310},
  doi={10.1109/IROS45743.2020.9341076}}

@inproceedings{Shirai:Pivoting:NeurIPS2025,
  author = {Shirai, Yuki and Ota, Kei and Jha, Devesh K. and Romeres, Diego},
  title = {{Sim-to-Real Contact-Rich Pivoting via Optimization-Guided RL with Vision and Touch}},
  booktitle = {NeurIPS 2025 Workshop on Embodied World Models for Decision Making},
  year = 2025,
}

@inproceedings{Lum:DextrAH-G:ACRL2024,
  title={{DextrAH-G}: Pixels-to-Action Dexterous Arm-Hand Grasping with Geometric Fabrics},
  author={Tyler Ga Wei Lum and Martin Matak and Viktor Makoviychuk and Ankur Handa and Arthur Allshire and Tucker Hermans and Nathan D. Ratliff and Karl Van Wyk},
  booktitle={Conference on Robot Learning (CoRL)},
  year={2024},
}

@article{Liu:FetchBot:2025,
      title={{FetchBot}: Learning Generalizable Object Fetching in Cluttered Scenes via Zero-Shot {Sim2Real}},
      author={Weiheng Liu and Yuxuan Wan and Jilong Wang and Yuxuan Kuang and Wenbo Cui and Xuesong Shi and Haoran Li and Dongbin Zhao and Zhizheng Zhang and He Wang},
      year={2025},
      journal={arXiv preprint arXiv:2502.17894},
}

@InProceedings{Ranawaka:SAIL:PMLR2025,
  title = 	 {{SAIL}: Faster-than-Demonstration Execution of Imitation Learning Policies},
  author =       {Arachchige, Nadun Ranawaka and Chen, Zhenyang and Jung, Wonsuhk and Shin, Woo Chul and Bansal, Rohan and Barroso, Pierre and He, Yu Hang and Lin, Yingyan Celine and Joffe, Benjamin and Kousik, Shreyas and Xu, Danfei},
  booktitle = 	 {Conference on Robot Learning (CoRL)},
  pages = 	 {721--749},
  year = 	 {2025},
  volume = 	 {305},
  publisher =    {PMLR}
}

@inproceedings{Tsai:DeepImageHarmonization:CVP2017,
  title = {Deep image harmonization},
  author = {Tsai, Yi-Hsuan and Shen, Xiaohui and Lin, Zhe and Sunkavalli, Kalyan and Lu, Xin and Yang, Ming-Hsuan},
  booktitle = {Proceedings of the IEEE Conference on Computer Vision and Pattern Recognition (CVPR)},
  pages = {2799--2807},
  year = {2017},
  doi = {10.1109/CVPR.2017.299},
}

@InProceedings{Xuanlin:EvalRealinSim:PMLR2025,
  title = 	 {Evaluating Real-World Robot Manipulation Policies in Simulation},
  author =       {Li, Xuanlin and Hsu, Kyle and Gu, Jiayuan and Mees, Oier and Pertsch, Karl and Walke, Homer Rich and Fu, Chuyuan and Lunawat, Ishikaa and Sieh, Isabel and Kirmani, Sean and Levine, Sergey and Wu, Jiajun and Finn, Chelsea and Su, Hao and Vuong, Quan and Xiao, Ted},
  booktitle = 	 {Conference on Robot Learning (CoRL)},
  pages = 	 {3705--3728},
  year = 	 {2025},
  volume = 	 {270},
  publisher =    {PMLR}
}

@article{Mandlekar:RoboMimic:2021,
  title={What Matters in Learning from Offline Human Demonstrations for Robot Manipulation},
  author={Ajay Mandlekar and Danfei Xu and Josiah Wong and Soroush Nasiriany and Chen Wang and Rohun Kulkarni and Li Fei-Fei and Silvio Savarese and Yuke Zhu and Roberto Mart\'{i}n-Mart\'{i}n},
  journal={arXiv preprint arXiv:2108.03298},
  year={2021}
}

@ARTICLE{Fabio:NonprehensileSurvey:RAL2018,
  author={Ruggiero, Fabio and Lippiello, Vincenzo and Siciliano, Bruno},
  journal={IEEE Robotics and Automation Letters}, 
  title={Nonprehensile Dynamic Manipulation: A Survey}, 
  year={2018},
  volume={3},
  number={3},
  pages={1711-1718},
  doi={10.1109/LRA.2018.2801939}
}

@article{Nikhila:SAM2:2024,
  title={{SAM 2}: Segment Anything in Images and Videos},
  author={Ravi, Nikhila and Gabeur, Valentin and Hu, Yuan-Ting and Hu, Ronghang and Ryali, Chaitanya and Ma, Tengyu and Khedr, Haitham and R{\"a}dle, Roman and Rolland, Chloe and Gustafson, Laura and Mintun, Eric and Pan, Junting and Alwala, Kalyan Vasudev and Carion, Nicolas and Wu, Chao-Yuan and Girshick, Ross and Doll{\'a}r, Piotr and Feichtenhofer, Christoph},
  journal={arXiv preprint arXiv:2408.00714},
  year={2024}
}

@article{Sergio:ArUcoMarker:PR2014,
  title={Automatic generation and detection of highly reliable fiducial markers under occlusion},
  author={Garrido-Jurado, Sergio and Mu{\~n}oz-Salinas, Rafael and Madrid-Cuevas, Francisco J and Mar{\'\i}n-Jim{\'e}nez, Manuel J},
  journal={Pattern Recognition},
  volume={47},
  number={6},
  pages={2280--2292},
  year={2014}
}

@INPROCEEDINGS{Josip:AnalysisRandomization:IROS2022,
  author={Josifovski, Josip and Malmir, Mohammadhossein and Klarmann, Noah and Žagar, Bare Luka and Navarro-Guerrero, Nicolás and Knoll, Alois},
  booktitle={Proceedings of the IEEE/RSJ International Conference on Intelligent Robots and Systems (IROS)}, 
  title={Analysis of Randomization Effects on Sim2Real Transfer in Reinforcement Learning for Robotic Manipulation Tasks}, 
  year={2022},
  volume={},
  number={},
  pages={10193-10200},
  doi={10.1109/IROS47612.2022.9981951}}

@INPROCEEDINGS{Josh:VisualDomainRandomization:IROS2017,
  author={Tobin, Josh and Fong, Rachel and Ray, Alex and Schneider, Jonas and Zaremba, Wojciech and Abbeel, Pieter},
  booktitle={Proceedings of the IEEE/RSJ International Conference on Intelligent Robots and Systems (IROS)}, 
  title={Domain randomization for transferring deep neural networks from simulation to the real world}, 
  year={2017},
  volume={},
  number={},
  pages={23-30},
  doi={10.1109/IROS.2017.8202133}}

@inproceedings{Haarnoja:SAC_Maniskill:PMLR2018,
  author       = {Tuomas Haarnoja and
                  Aurick Zhou and
                  Pieter Abbeel and
                  Sergey Levine},
  title        = {Soft Actor-Critic: Off-Policy Maximum Entropy Deep Reinforcement Learning
                  with a Stochastic Actor},
  booktitle    = {Proceedings of the 35th International Conference on Machine Learning ({ICML})},
  volume       = {80},
  pages        = {1856--1865},
  publisher    = {{PMLR}},
  year         = {2018},
  bibsource    = {dblp computer science bibliography, https://dblp.org}
}

@article{Schulman:PPO:arXiv2017,
  title={Proximal policy optimization algorithms},
  author={Schulman, John and Wolski, Filip and Dhariwal, Prafulla and Radford, Alec and Klimov, Oleg},
  journal={arXiv preprint arXiv:1707.06347},
  year={2017}
}

@inproceedings{Zhao:ACT:RSS2023,
  title        = {Learning Fine-Grained Bimanual Manipulation with Low-Cost Hardware},
  author       = {Zhao, Tony Z. and Jain, Ajay and Ichter, Brian and Levine, Sergey},
  booktitle    = {Proceedings of Robotics: Science and Systems (RSS)},
  year         = {2023}
}

@inproceedings{Chi:DP:RSS2023,
  title        = {Diffusion Policy: Visuomotor Policy Learning via Action Diffusion},
  author       = {Chi, Cheng and Feng, Siyuan and Du, Yilun and Xu, Zichen and Cousineau, Etienne and Burchfiel, Benjamin and Song, Shuran},
  booktitle    = {Proceedings of Robotics: Science and Systems (RSS)},
  year         = {2023}
}

@inproceedings{Chen:ClutterDexGrasp:CoRL2025,
  title     = {{ClutterDexGrasp}: A Sim-to-Real System for General Dexterous Grasping in Cluttered Scenes},
  author    = {Chen, Zeyuan and Yan, Qiyang and Chen, Yuanpei and Wu, Tianhao and Zhang, Jiyao and Ding, Zihan and Li, Jinzhou and Yang, Yaodong and Dong, Hao},
  booktitle = {Conference on Robot Learning (CoRL)},
  year      = {2025}
}

@inproceedings{Yarats:DrQ-v2:ICLR2022,
  title={Mastering Visual Continuous Control: Improved Data-Augmented Reinforcement Learning},
  author={Denis Yarats and Rob Fergus and Alessandro Lazaric and Lerrel Pinto},
  booktitle={International Conference on Learning Representations},
  year={2022},
}

@inproceedings{Siyunan:Maniskill3:RSS2025,
  title        = {{ManiSkill3}: {GPU} Parallelized Robotics Simulation and Rendering for Generalizable Embodied {AI}},
  author       = {Tao, Siyuan and Xiang, Fanbo and Shukla, Aayush and Qin, Yijiong and Hinrichsen, Xavier and Yuan, Chunrui and Bao, Chen and Lin, Xupeng and Liu, Yifan and Chan, Tony-Kit and Gao, Yunchang and Li, Xiang and Mu, Tongzhou and Xiao, Ning and Gurha, Ashish and Rajesh, Venkatesh Narayan and Choi, Yeo Won and Chen, Yu-Ren and Huang, Zeyu and Calandra, Roberto and Chen, Ruoxi and Luo, Siyuan and Su, Hao},
  booktitle    = {Proceedings of Robotics: Science and Systems (RSS)},
  year         = {2025}
}

@ARTICLE{Sorokin:Sidewalk:RAL2022,
  author  = {Maks Sorokin and Jie Tan and C. Karen Liu and Sehoon Ha},
  title   = {Learning to Navigate Sidewalks in Outdoor Environments},
  journal = {IEEE Robotics and Automation Letters},
  volume  = {7},
  number  = {2},
  pages   = {3906--3913},
  year    = {2022},
  doi     = {10.1109/LRA.2022.3145947}
}

@ARTICLE{Singh:DextrAH-RGB:2024,
  author  = {Ritvik Singh and Arthur Allshire and Ankur Handa and Nathan Ratliff and Karl Van Wyk},
  title   = {{DextrAH-RGB}: Visuomotor Policies to Grasp Anything with Dexterous Hands},
  journal = {arXiv preprint arXiv:2412.01791},
  year    = {2024}
}

@INPROCEEDINGS{He:VIRAL:CVPR2026,
  author    = {Tairan He and Zi Wang and Haoru Xue and Qingwei Ben and Zhengyi Luo and Wenli Xiao and Ye Yuan and Xingye Da and Fernando Casta{\~n}eda and Shankar Sastry and Changliu Liu and Guanya Shi and Linxi Fan and Yuke Zhu},
  title     = {{VIRAL}: Visual Sim-to-Real at Scale for Humanoid Loco-Manipulation},
  booktitle = {Proceedings of the IEEE/CVF Conference on Computer Vision and Pattern Recognition (CVPR)},
  year      = {2026}
}

@ARTICLE{Chen:PTLD:2026,
  author  = {Rosy Chen and Mustafa Mukadam and Michael Kaess and Tingfan Wu and Francois R. Hogan and Jitendra Malik and Akash Sharma},
  title   = {{PTLD}: Sim-to-Real Privileged Tactile Latent Distillation for Dexterous Manipulation},
  journal = {arXiv preprint arXiv:2603.04531},
  year    = {2026}
}

@ARTICLE{Kase:MotorBabbling:JRM2021,
  author  = {Kei Kase and Noboru Matsumoto and Tetsuya Ogata},
  title   = {Leveraging Motor Babbling for Efficient Robot Learning},
  journal = {J. Robot. Mechatron.},
  volume  = {33},
  number  = {5},
  pages   = {1063--1074},
  year    = {2021},
  doi     = {10.20965/jrm.2021.p1063}
}

@ARTICLE{Sueoka:RealWorldCooperativeTransport:JRM2025,
  author  = {Yuichiro Sueoka and Seiya Ozaki and Yuki Kato and Takahiro Yoshida},
  title   = {Real-World Validation of a Learned Policy for Cooperative Object Transport with the Transformer Encoder},
  journal = {J. Robot. Mechatron.},
  volume  = {37},
  number  = {5},
  pages   = {1034--1041},
  year    = {2025},
  doi     = {10.20965/jrm.2025.p1034}
}

@ARTICLE{Kawarai:WholeArmManipulation:JRM2010,
  author  = {Nobuyuki Kawarai and Yuichi Kobayashi},
  title   = {Learning of Whole Arm Manipulation with Constraint of Contact Mode Maintaining},
  journal = {J. Robot. Mechatron.},
  volume  = {22},
  number  = {4},
  pages   = {542--550},
  year    = {2010},
  doi     = {10.20965/jrm.2010.p0542}
}

\end{document}